\documentclass[conference]{IEEEtran}
\IEEEoverridecommandlockouts
\usepackage{cite}
\usepackage{graphicx}
\usepackage{textcomp}
\usepackage{xcolor}
\usepackage[nolist]{acronym}
\usepackage{algorithm}
\usepackage{dsfont}
\usepackage{algpseudocode}
\usepackage{multirow}
\usepackage{siunitx}
\usepackage{booktabs}
\usepackage{float}
\usepackage{subfigure}
\usepackage{comment}

\def\BibTeX{{\rm B\kern-.05em{\sc i\kern-.025em b}\kern-.08em
    T\kern-.1667em\lower.7ex\hbox{E}\kern-.125emX}}

\usepackage[inline]{enumitem}

\usepackage[absolute,showboxes]{textpos}

\TPMargin{5pt}
  
\newcommand{\copyrightstatement}{
    \begin{textblock}{0.84}(0.08,0.01)    
         \noindent
         \footnotesize
         \copyright 2024 IEEE. Personal use of this material is permitted. Permission from IEEE must be obtained for all other uses, in any current or future media, including reprinting/republishing this material for advertising or promotional purposes, creating new collective works, for resale or redistribution to servers or lists, or reuse of any copyrighted component of this work in other works.
    \end{textblock}
}

\begin{document}

\copyrightstatement

\title{Feature Distribution Shift Mitigation with Contrastive Pretraining for Intrusion Detection

\thanks{This work has received funding from the German Research Fundation (DFG) under grant numbers MA 6529/4-1 and KE 1863/10-1.}
}

\author{\IEEEauthorblockN{Weixing Wang\textsuperscript{*\textdagger}, Haojin Yang\textsuperscript{*}, Christoph Meinel\textsuperscript{*},\\Hasan Yagiz {\"O}zkan\textsuperscript{\textdagger}, Cristian Bermudez Serna\textsuperscript{\textdagger}, Carmen Mas-Machuca\textsuperscript{\textdaggerdbl}\textsuperscript{\textdagger}}
\IEEEauthorblockA{\textsuperscript{*}Hasso Plattner Institut (HPI), Germany}
\IEEEauthorblockA{\textsuperscript{\textdagger}Chair of Communication Networks (LKN), Technical University of Munich (TUM), Germany}
\IEEEauthorblockA{\textsuperscript{\textdaggerdbl}Chair of Communication Networks, University of the Bundeswehr Munich, Germany}\{weixing.wang, haojin.yang, christoph.meinel\}@hpi.de, \{yagiz.oezkan, cristian.bermudez-serna\}@tum.de, cmas@unibw.de}

\maketitle

\begin{abstract}
In recent years, there has been a growing interest in using \acf{ML}, especially \acf{DL} to solve \acf{NID} problems. However, the feature distribution shift problem remains a difficulty, because the change in features' distributions over time negatively impacts the model's performance. As one promising solution, model pretraining has emerged as a novel training paradigm, which brings robustness against feature distribution shift and has proven to be successful in \acf{CV} and \acf{NLP}. To verify whether this paradigm is beneficial for \ac{NID} problem, we propose SwapCon, a \ac{ML} model in the context of \ac{NID}, which compresses shift-invariant feature information during the pretraining stage and refines during the finetuning stage. We exemplify the evidence of feature distribution shift using the Kyoto2006+ dataset. We demonstrate how pretraining a model with the proper size can increase robustness against feature distribution shifts by over 8\%. Moreover, we show how an adequate numerical embedding strategy also enhances the performance of pretrained models. Further experiments show that the proposed SwapCon model also outperforms \ac{XGBoost} and \ac{KNN} based models by a large margin.
\end{abstract}


\section{Introduction}

A \ac{NID} system monitors and analyzes network traffic to identify and respond to unauthorized or malicious activities. These systems play a vital role in protecting modern networks from threats such as malware, ransomware, and other cyber-attacks \cite{nidef}. According to the 2022 Cyber-threat Defense Report \cite{cyberedge2022}, the number of organizations that experienced at least one successful cyber-attack increased from 61.9\% to 81.3\% since 2014, and the mean annual IT security budget increased by 4.6\% in the year 2022. Failing to secure a network properly can have severe financial and reputational consequences. For example, the 2017 Equifax data breach, which exposed the personal data of 147 million people, resulted in a loss of \$439 million for the company.

\ac{ML} algorithms can quickly and accurately identify patterns in large amounts of data. This ability makes \ac{ML} models well suited for detecting anomalies, which may indicate intrusions in a network. However, applying \ac{ML} for \ac{NID} comes with the challenge of the feature distribution shift. For example, consider an \ac{ML} model that is trained to detect network intrusions using data collected during 2005 and is later deployed in the same network in 2006. The model may not perform as accurately as it did during training due to the change in network configurations and user behaviors, which leads to a shift in the distribution of the features in the dataset that the \ac{ML} model has not seen during training. 

Model pertaining is a promising method to mitigate the aforementioned problem. In pretraining, an \ac{ML} model is trained on a large dataset in an unsupervised manner to learn general patterns, which can be later finetuned for specific tasks. For instance, Hendrycks et al.~\cite{robustden} demonstrated how pretraining can increase model robustness against feature distribution shift by approximately 10\% in image classification tasks.

The main contributions of this work are: \textit{i)} we introduce SwapCon, a pretrained \ac{ML} model robust against the feature distribution shift for \ac{NID}. \textit{ii)} We leverage contrastive pretraining for finding pattern representations in the data and a swapping augmentation strategy for creating positive and negative samples. Moreover, we use different embedding methods for numerical and categorical features to increase the feature space. \textit{iii)} We test our model on the open-source Kyoto2006+ dataset \cite{kyoto2006}, which contains 10 years of network traffic information. And \textit{iv)} we study the effect of pretraining using the SwapCon model, while comparing our best variant with a \ac{KNN} and an \ac{XGBoost} based models. The results show that model pretraining increases robustness against the feature distribution shift in the selected dataset, while the other models suffer from performance drops.


\section{Preliminary}
\label{sec:preliminary}

\subsection{Feature Distribution Shift}
\label{sec:ds}
Feature Distribution Shift refers to a change in the distribution features between the training and testing sets. It can be caused by various factors, such as changes in user behavior, technological advancements, and shifts in popular applications or services.



\subsection{Contrastive Learning}
\label{sec:contra}
\ac{CL} aims to find a proper latent space for the original feature space. In this space, similar samples have a closer distance than dissimilar samples. To this end, a contrastive objective function is needed. In this work, the \ac{NCE} loss function is applied. It converts the problem of modeling complex probability distributions of two feature spaces into a simpler binary classification task, where the model discriminates between genuine data samples and noisy samples, making training more computationally efficient and scalable.

\subsection{Model Pretraining}
\label{sec:pretrain}

\begin{figure}[htp]
    
    
    \centering
    \includegraphics[width=\columnwidth]{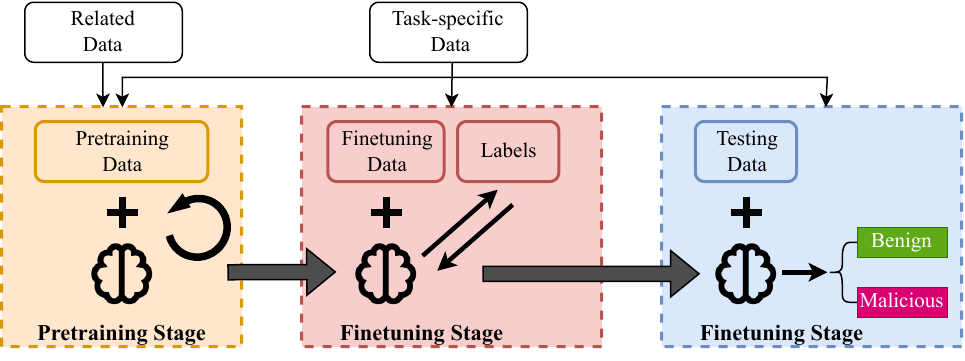}
    
    
    \caption{\ac{ML} model training pipeline with pretraining.}
    \label{fig:pretrainexp}
\end{figure}

Pretrained models have achieved massive success in the field of \ac{CV} as well as \ac{NLP} and have become the new paradigm for many \ac{ML} tasks \cite{pretrainsuvey}. Model pretraining, as depicted in Fig.~\ref{fig:pretrainexp}, adds an initial stage in the model training pipeline, where the model is pretrained on related datasets. Those datasets should be large and diverse, covering a wide range of variations, patterns, and examples relevant to the target task. For example, the famous \acs{BERT} model is pretrained using many text sources such as English Wikipedia. The goal in the pretraining stage is to capture as much knowledge as possible from the data and store it in the model. After pretraining, the model is finetuned on the task-specific dataset. The knowledge gained during pretraining enables the model to efficiently learn the target task. For example, a pretrained language model can be finetuned for either sentiment analysis, language translation, or sentence classification. 

In tabular data, each piece of sample is a row containing a series of features. For this data representation, it is also possible to perform \ac{CL}. Somepalli et al. \cite{saint} proposed a pretraining framework that first applies augmentation operations on rows and trains a transformer-based encoder using a contrastive objective function. Positive pairs are generated by mixing up several rows in a batch and the \ac{NCE} loss function is used to minimize the distance between positive samples while distancing negative samples. In this work, contrastive pretraining is applied with a different data augmentation method.

\subsection{\ac{ML} Models for the \ac{NID} Problem}
\acp{DT} are one of the basic supervised \ac{ML} algorithms used for classification tasks. Despite their simplicity, \ac{DT} models show superior performance in many tasks, including \ac{NID}. XGBoost is a \ac{DT} model designed for gradient-boosting. The key idea about gradient boosting is to assemble multiple single \ac{DT}s to boost the model performance. In \cite{xgboostids}, Dhaliwal et al. applied the XGBoost model on the NSL-KDD dataset and reached a 98.7\% accuracy score on the test dataset. To the best of our knowledge, \cite{anoshift} is the first and so far, the only work that studies the feature distribution shift using a \ac{NID} dataset, where the researchers proposed both \ac{ML} and \ac{DL} models. 

\section{Problem}
\label{sec:problem}

\begin{figure}[h!]
     
    \centering
    \includegraphics[clip,width=0.6\linewidth]{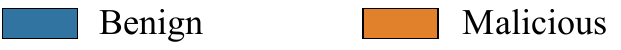}
    \includegraphics[clip,width=\linewidth]{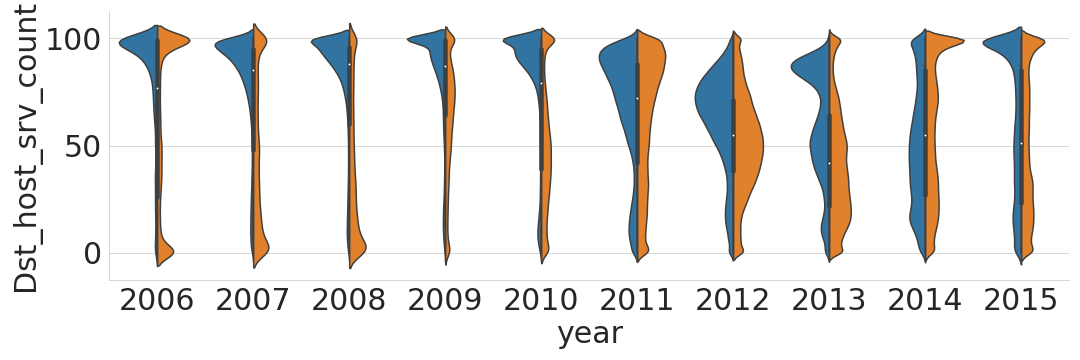}
    \caption{Visualization of the shapes of a feature distribution shift over time in the Kyoto2006+ dataset~\cite{kyoto2006}. In each year, the horizontal expansion of the feature plot shows its probability density. The Y-axis means the percentage value that ranges from 0 to 100 which relates to the feature values.}
    \label{fig:feadst}

\end{figure}

Using the same visualization method of Dr{\u{a}}goi et al. \cite{anoshift}, we show the evidence of feature distribution shift over time in the Kyoto2006+ dataset. In Fig.~\ref{fig:feadst}, the distribution of the network traffic feature \textit{Dst\_host\_srv\_count} is visualized on a yearly basis. As we can see, the feature distributions are similar within the first five years. Then, there is a sudden change starting from the year 2011.

Although the feature distribution shift is often present, it is likewise overseen. Song et al. demonstrated that the feature distributions of network traffic in the Kyoto2006+ dataset vary even from month to month \cite{kyoto2006}. Ignoring the feature distribution shift during model training can lead to serious consequences in the testing or deployment stages. Koh et al. constructed a benchmark containing datasets with different types of feature distribution shifts and demonstrated how standard training strategies yield substantially worse results in those datasets \cite{koh2021wilds}.

One way of mitigating the distribution shift problem is to continuously update and retrain \ac{ML} models using the most recent and representative data. This allows the models to adapt to the changing distributions and maintain their accuracy in classifying network traffic. However, the cost of training models and labeling new data is often too large to be practical.


\section{Design}
\label{sec:design}

\subsection{Dataset}
The Kyoto2006+ dataset \cite{kyoto2006}, which was built on 10 years of real network traffic data from 2006 to 2015, is used in this work. It consists 24 features and two classes which are denoted as benign and malicious. In our study, 14 out of 24 features are used to train \ac{ML} models since the other 10 features are not generic. We refer the reader to \cite{kyoto2006} for more details. 

\begin{figure}[htp]

    \centering
    \includegraphics[width=\columnwidth]{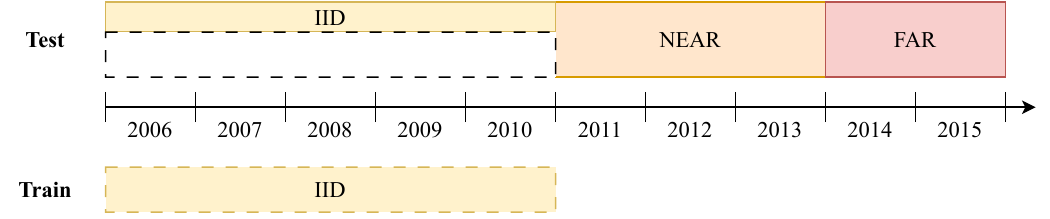}

    \caption{Splits of \textit{IID}, \textit{NEAR}, and \textit{FAR} based on 10 years of data. Note that the training set is sampled from the \textit{IID} split.}
    \label{fig:splits}
\end{figure}

We follow the proposed chronological approach in \cite{anoshift} to split all 10 years of data into three main splits that can highlight the feature distribution shift over time. Fig.~\ref{fig:splits} shows the three splits on a time axis as well as the train/test split.

\textit{IID},  \textit{NEAR} and \textit{FAR} represent the different degrees of feature distribution shift. We denote the first split as \textit{IID} (Independent and Identically Distributed) because the data distribution is the same as the training set. We assume that more recent data with respect to the \textit{IID} split should have less feature distribution shift than more distant data. Models are pretrained and finetuned within the training set and tested in the three splits separately. The training set is divided into pretraining and fine-tuning sets with a ratio of 9:1, respectively. To remove the bias in the testing stage, 5000 benign and 5000 malicious samples are sampled from the original testing set, making in total of 10000 for each testing set.

\subsection{SwapCon Overview}

SwapCon consists of a stack of blocks where each block is a combination of a linear layer followed by a \ac{BN} layer. \ac{BN} normalizes the layer activations in a neural network within each batch during training. By reducing the internal covariate shift, which is the change in the distribution of activations across different layers during training, \ac{BN} facilitates faster learning and helps the network reach a better optimum. In SwapCon, two kinds of activation functions are applied after linear layers. The \textit{Tanh} is used in the first layer and the \textit{ReLu} activation function is applied in the rest of the layers.

\begin{figure*}[h!]
     \centering
     \begin{minipage}[b]{.52\textwidth}
         \centering
         \includegraphics[width=\textwidth]{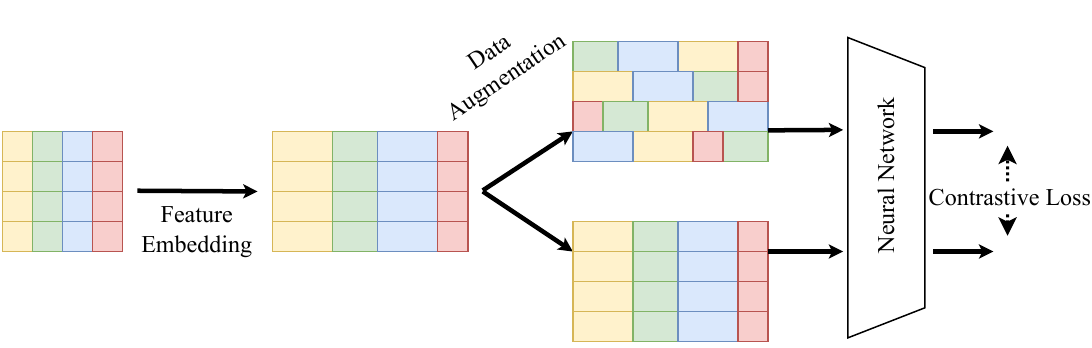}
         \caption{SwapCon pretraining pipeline. Colors represent different features.}
         \label{fig:pretrain}
     \end{minipage}
     \quad
     \begin{minipage}[b]{.45\textwidth}
         \centering
         \includegraphics[width=\textwidth]{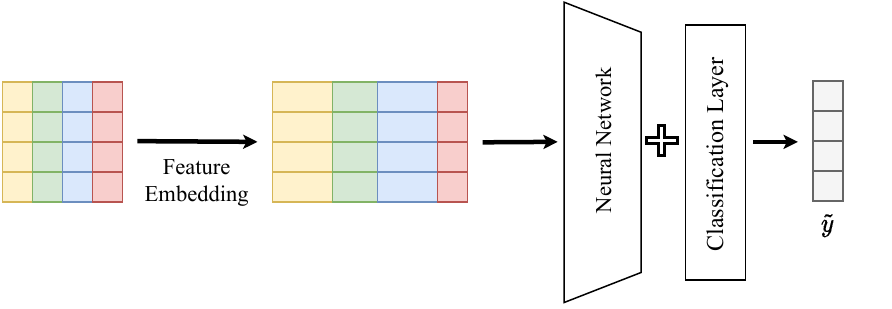}
         \caption{The finetuning pipeline of SwapCon.}
         \label{fig:finetune}
     \end{minipage}
        \label{fig:Swapcon}
\end{figure*}


\subsection{Contrastive Pretraining}
\label{sec:swapconpretrain}
In pretraining, \ac{CL} is leveraged to inject invariant latent features in the model weights in a self-supervised way. Fig.~\ref{fig:pretrain} depicts the high-level pipeline of the SwapCon pretraining procedure. The input is a mini-batch of samples which are represented as rows in the figure. In the first step, the feature embedding operation embeds the numerical or categorical feature into a higher dimension. After the data augmentation operation, both the original and augmented batches are processed by the same neural network. The \ac{NCE} loss function in Eq.~\ref{eq:infonce} is applied to the outputs. 

\ac{CL} relies on positive and negative sample pairs, which requires finding at least one negative and one positive variant of a single sample. While it is easy to regard other samples in the same mini-batch as negative ones, data augmentation techniques are applied to generate positive ones. In SwapCon, sample-based data augmentation is applied by randomizing the order of features in a sample. The augmented sample preserves all original information and thus is regarded as a positive instance. By comparing the contrastive pairs, the model learns the order-invariant characteristic of tabular data. This method yields a simple and effective self-supervised task.

\begin{equation}
\label{eq:infonce}
    \mathcal{L}_{NCE}^{(i, j)}=-\log \frac{\exp \left(\textit{sim}\left(\textbf{h}_i, \textbf{h}_j\right) / \tau\right)}{\sum_{k=1}^{2 N} \mathds{1}_{[k \neq i]} \exp \left(\textit{sim}\left(\textbf{h}_i, \textbf{h}_k\right) / \tau\right)}
\end{equation}
In pretraining, the \ac{NCE} loss function of Eq.~\ref{eq:infonce} is considered, where $\mathds{1}_{[k \neq i]}$ is the identity function and $\tau$ is a hyperparameter that controls the importance of negative samples. For simplicity, we use one as the default $\tau$ value. $\textbf{h}_i$ and $\textbf{h}_j$ represent the original sample and its augmented version, respectively. The similarity function \textit{sim} computes the cosine similarity between the two representations. Given an anchor sample $h_i$, the model tries to distinguish the positive from a set of negative samples by minimizing the $\mathcal{L}_{NCE}^{(i, j)}$ function.

\subsection{Finetuning Strategy}
\label{sec:finetunestrategy}

After the pretraining stage, the model is finetuned with supervised learning. Fig.~\ref{fig:finetune} depicts this procedure. Here, additional classification layers are stacked after the pretrained model for the binary classification task.

There are two ways for finetuning. The first one is called full finetuning, where the weights of both the pretrained model and the classification layer are updated during training. The second is called partial finetuning, where only the weights of the last classification layers are updated and the weights in the pretrained model are fixed. While the first method leads to faster convergence and possibly higher performance, it breaks the data invariance information learned during the pretraining stage. Both methods are applied and discussed in Section~\ref{sec:evaluation}. 


\subsection{Numerical Feature Embedding} 
\label{section:num}
Guo et al. demonstrated that numerical feature embedding often brings better performance for \ac{ML} models \cite{numemb}. Hence, three numerical feature embedding strategies are implemented and studied. The first two methods are variants of the binning method, which is used to categorize a range of varied, continuous values into a smaller number of distinct bins.
\subsubsection{\ac{EB}} 
\begin{figure}[H]
  \begin{center}
    \includegraphics[width=0.8\linewidth]{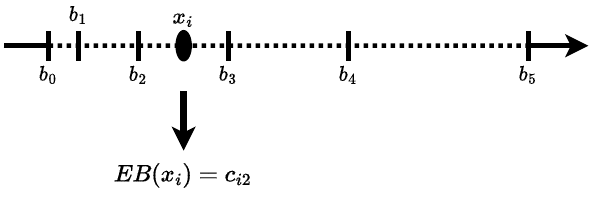}
    \caption{Illustration of the \ac{EB} strategy. Here $x_i$ is the i-th feature in the sample. }
    \label{fig:binning}
  \end{center}
\end{figure}

This is used when the data is concentrated at one end of the value range. As such, bins have finer intervals where the numerical values are also denser. As depicted in Fig.~\ref{fig:binning}, the solid line represented the range of a numerical feature in the training set and a bin is marked within $[b_t,b_{t+1}]$. Every numerical feature that has the value covered by a bin will be assigned to the corresponding category. This is also the baseline embedding method used in our experiments.

\subsubsection{\ac{PLE}} 
\begin{figure}[h!]
  \begin{center}
    \includegraphics[width=0.8\linewidth]{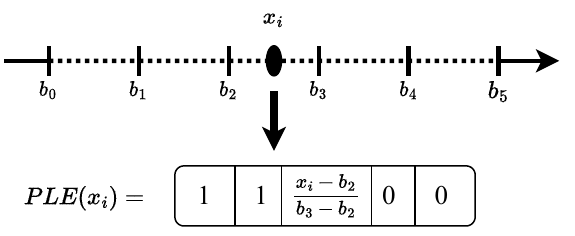}
    \caption{Illustration of the \ac{PLE} strategy. }
    \label{fig:ple}
  \end{center}
\end{figure}

Proposed by Gorishniy et al. \cite{numemb1} is inspired by the one-hot embedding method with a more precise representation of the numerical value. In \ac{PLE}, each feature value range is split into a disjoint set of bins with the same width. A numerical feature is represented with a vector of length $T$, where $T$ is the number of bins. We set $T$ to be 128 in all experiments. Fig.~\ref{fig:ple} depicts the \ac{PLE} encoding strategy.

PLE offers multiple benefits. As a preprocessing strategy, \acs{PLE} provides a more fine-grained numerical value representation. Normally, a part of the information is lost when a scalar number is categorized into bins, but \acs{PLE} establishes a direct bijection between the single dimension and a higher dimension, which enlarges the feature space for the following learning process. Moreover, \acs{PLE} preserves the inherent magnitude relationship of the numbers. In the \ac{EB} method, numbers are converted into categories, thus losing their magnitude and can not be compared. But with \ac{PLE}, two numbers can still be compared after the embedding.

\subsubsection{\ac{LE}}
\begin{figure}[h!]
  \begin{center}
    \includegraphics[width=0.4\linewidth]{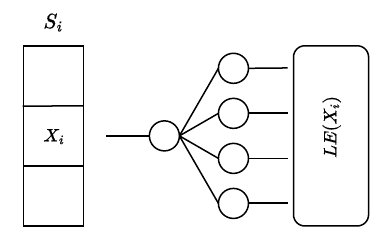}
    \caption{Illustration of the \ac{LE} strategy. The size of the embedding is controlled by the linear layer. }
    \label{fig:LE}
  \end{center}
\end{figure}

It is a model-aware embedding schema, where an encoder for each numerical feature is jointly trained with the model in the pretraining stage. Each numerical feature of a sample is fed into a linear layer that does not share weights with others. After pretraining, the encoders are fixed. In \ac{LE}, the same values of different features should have different meanings and thus different representations. Fig.~\ref{fig:LE} shows how \ac{LE} works.

\section{Evaluation}

\label{sec:evaluation}

 \subsection{Pretraining with SwapCon}
 

In the first experiment, SwapCon is pretrained with a network with three layers, and another three classification layers are added during the finetuning stage. The model is pretrained with two epochs and a mini-batch with 512 samples. Note that all experiments are averaged over three executions.

\begin{table}[h!]
  \centering
  \caption{Performance of model with 6 layers.}
  \begin{tabular}{l|S|S|S|S}
    \toprule
    \multirow{2}{*}{AUC($\uparrow$)} &
      \multicolumn{2}{c|}{With Pretrain} & \multicolumn{2}{c}{Without Pretrain} \\ \cmidrule{2-5}&
      {Fix} & {No-Fix} & {Fix} & {No-Fix} \\
      \midrule\midrule
    IID& 98.83 & \textbf{99.28} & 97.58 & 99.12 \\
    NEAR& 94.25 & \textbf{98.81} & 94.01 & 97.23 \\
    FAR& 67.92 & 42.71 & \textbf{68.81} & 42.08 \\
    \bottomrule
  \end{tabular}
  \label{tab:swap1}
\end{table}

\subsection{Finetuning with SwapCon}

\begin{table*}[h!]
\centering
\caption{Model performance when fixing from 0 to all layers.}
\label{tab:swap2}
\begin{tabular}{l|lllllll}
\toprule
AUC($\uparrow$) & No-Fix         & \multicolumn{1}{c}{\begin{tabular}[c]{@{}c@{}}Fix 1. \\ Layer\end{tabular}} & \multicolumn{1}{c}{\begin{tabular}[c]{@{}c@{}}Fix 1$\sim$2 \\ Layers\end{tabular}} & \multicolumn{1}{c}{\begin{tabular}[c]{@{}c@{}}Fix 1$\sim$3 \\ Layers\end{tabular}} & \multicolumn{1}{c}{\begin{tabular}[c]{@{}c@{}}Fix 1$\sim$4 \\ Layers\end{tabular}} & \multicolumn{1}{c}{\begin{tabular}[c]{@{}c@{}}Fix 1$\sim$5 \\ Layers\end{tabular}} & \multicolumn{1}{c}{\begin{tabular}[c]{@{}c@{}}Fix 1$\sim$6 \\ Layers\end{tabular}} \\ \midrule\midrule
IID   & \textbf{99.12} & 98.31                                                                       & 98.33                                                                              & 97.93                                                                              & 96.77                                                                              & 91.31                                                                              & 49.76                                                                              \\
NEAR  & 97.32          & \textbf{97.68}                                                              & 97.63                                                                              & 95.27                                                                              & 92.67                                                                              & 86.77                                                                              & 51.59                                                                              \\
FAR   & 42.08          & 50.28                                                                       & 52.49                                                                              & 64.89                                                                              & \textbf{65.84}                                                                     & 65.07                                                                              & 53.63                                                                             \\ \bottomrule
\end{tabular}
\end{table*}

In the finetune stage, the early-stopping strategy is adopted, which means the training procedure will be stopped when the performance on the validation set starts to deteriorate. With this strategy, all training finishes within two epochs. For simplicity, we use the \ac{AUC} metric on a balanced testing set.

As shown in Tbl.~\ref{tab:swap1}, four variants of models are presented.  Note that \textit{Fix} means partial finetuning and \textit{No-Fix} means full finetuning. Models without pretraining would have random initial parameter weights for the first three layers, while models with pretraining inherit weights from that stage. Here, partial finetuning means fixing the weights from the first three layers and only training the classification layers during finetuning. The best results in each split are denoted in bold. By comparing the table rows, we see that the performances of all variants decrease as the time distance becomes larger. Comparing against the results in the \textit{IID} split, the performance drops from 0.5\% to 4\% in the \textit{NEAR} split and up to 50\% in the \textit{FAR} split. This demonstrates that feature distribution shifts can substantially undermine model performance. A larger gap in feature distributions means a larger drop in performance. However, the next section shows that a more sophisticated pretraining strategy can mitigate the problem to some extent.

\subsubsection{Effect of Pretraining}
The difference between pretrained and non-pretrained models lies in the first layers. Those pretrained layers are expected to deliver better data representations which could facilitate finetuning. By comparing column one with column three and column two with column four in Tbl.~\ref{tab:swap1}, we could see that the model performance increases for the \textit{IID} and \textit{NEAR} splits when using pretrained weights. It should not be surprising for the \textit{IID} split because the pretraining is also done using the data from this split. We could interpret the gain in performance as a result of training on more data. However, thanks to the knowledge learned through the unsupervised pretraining, the model performance increases by around 1\% in the \textit{NEAR} split. Nevertheless, pretraining seems to have side effects in the \textit{FAR} split. A pretrained model does not necessarily perform better than the model trained from scratch. This may be because the knowledge learned in the pretraining stage becomes less representative when the feature distribution gap becomes large. 

\subsubsection{Effect of Fixing Model Weights}
\label{par:modelsize}
By looking at Tbl.~\ref{tab:swap1} again, we could also find whether fixing the model weights also has different effects on the three splits. This time we compare column one with column two and column three with column four. For the \textit{IID} and \textit{NEAR} splits, there is a slight increase when the first layers are not fixed, and for the \textit{FAR} split there is a huge decrease of over 25\%. When fixing the first layers, the trainable parameters become less, which is equivalent to having a simpler model. And since the model is trained on the \textit{IID} split, the performance drop in this split is understandable. The \textit{NEAR} split is very close to the \textit{IID} split and the same effect applies. However, since the distributions of the features in the \textit{FAR} split are very different from those in the other two sets, a more complex model that learns too "well" on the \textit{IID} split suffers from the feature distribution shift the most. In contrast, a simpler model cannot be optimized as precisely as a complex model in the \textit{IID} split, but it gains some robustness against dissimilar data in the \textit{FAR} split.


\subsection{Determining the Best Model Size}

From the previous results, we found that although larger models can perform better in the \textit{IID} and \textit{NEAR} splits, they generalize worse in the \textit{FAR} splits. So we trained models with different numbers of fixed layers to verify the hypothesis between the model size and their generalization ability. As shown in Tbl.~\ref{tab:swap2}, as more layers are fixed, the model performance decreases in the \textit{IID} and \textit{NEAR} split. In the \textit{FAR} split, the model performs better in the beginning but then worsens in the end. We further notice that when using only three layers, the model performs well enough against feature distribution shift in the \textit{FAR} split and preserves the most performance in the other two splits.

Combining the above analysis, we train a model with three layers. The first two layers are pretrained and the last layer is the classification layer which is finetuned. Tbl.~\ref{tab:swap3} includes all variants. In the first row, SwapCon(3+3) means that the models have six layers, in which the first three layers are pretrained. Likewise, SwapCon(2+1) means that out of a total three layers, the first two of them are used for pretraining. The best combination is marked with *.

\begin{table*}[h]
  \centering
  \caption{All variants of the model trained with SwapCon.}
  \label{tab:swap3}
  \begin{tabular}{l|S|S|S|S|S|S|S|S}
    \toprule
    \multirow{4}{*}{AUC($\uparrow$)} &
      \multicolumn{4}{c|}{SwapCon (3+3)} & \multicolumn{4}{c}{SwapCon (2+1)$^*$}  \\ 
      \cmidrule{2-9} &
      \multicolumn{2}{c|}{With Pretrain} & \multicolumn{2}{c|}{Without Pretrain} & \multicolumn{2}{c|}{With Pretrain$^*$} & \multicolumn{2}{c}{Without Pretrain}\\ 
      \cmidrule{2-9}&
      {Fix} & {No-Fix} & {Fix} & {No-Fix} & {Fix$^*$} & {No-Fix} & {Fix} & {No-Fix} \\
      \midrule\midrule
    IID& 98.83 & \textbf{99.28} & 97.58 & 99.12 &98.64 & 98.99&94.82 & 97.53\\
    NEAR& 94.25 & \textbf{98.81} & 94.01 & 97.23 &98.32 &98.12 &89.63 & 95.14\\
    FAR& 67.92 & 42.71 & 68.81 & 42.08 &\textbf{72.31} & 70.99& 64.02& 64,78\\
    \bottomrule
  \end{tabular}
  
\end{table*}

By comparing column seven with column eight and column five with column six in Tbl.~\ref{tab:swap3}, we find that fixing the first two layers without pertaining will decrease the model performance since only one layer can be trained. At the same time, when the first two layers are pretrained, fixing them would bring a performance boost, which addresses the benefits of pretraining.

\subsection{Better Numerical Embedding}
\label{sec:numemb}
As demonstrated by Guo et al. in \cite{numemb}, the importance of numerical feature embedding is often overlooked. When looking at the data, we notice that different features may have the same numerical values, but the meanings behind the numbers differ. Thus various features should have different embedding vectors in a latent feature space. To bring scalar features into higher dimensions, we adopt the three previously introduced numerical feature embedding strategies and evaluate their impacts on the pretrained models. Notice that the \ac{EB} method is the baseline that is used in the previous experiments.

\begin{table}[h]
  \centering
  \caption{Impact of different numerical embedding techniques.}
  \begin{tabular}{l|S[table-column-width=2cm]|S[table-column-width=2cm]|S[table-column-width=2cm]}
    \toprule
    \multirow{2}{*}{AUC($\uparrow$)}
      & \multicolumn{3}{c}{SwapCon (2+1) \& With Pretrain \& Fix}  \\ 
      \cmidrule{2-4}&
       {\ac{EB}} & {\ac{PLE}} & {\ac{LE}}\\
      \midrule\midrule
    IID& 98.64 & 99.35 & \textbf{99.65}\\
    NEAR& 98.32 & 98.85 & \textbf{98.94}\\
    FAR& 72.31 & \textbf{72.93} & 70.99\\
    \bottomrule
  \end{tabular}
  \label{tab:numemb}
\end{table}

Tbl.~\ref{tab:numemb} shows the results on selected models. Note that for simplicity we only use the best model from the previous section. As we can see, the \acs{PLE} method offers a 0.5\%$\sim$0.7\% percent performance increase for all three splits. The \acs{LE} method delivers the best results in \textit{IID} and \textit{NEAR} splits; however, it decreases the model performance in the \textit{FAR} split. The reason may be the mismatch between feature representations and the actual samples. When pretraining with the \ac{LE} method, we essentially train a series of 1-layer neurons for numerical feature representation only considering the data in the \textit{IID} split. When we test the model in the \textit{IID} and \textit{NEAR} splits, the fixed \ac{LE} components can provide the most suitable embeddings so that the final results are better than the other methods. But this way of embedding numerical features becomes an unwanted bias in the \textit{FAR} split and leads to the worst performance. This experiment shows that embeddings should be contextualized and updated for new data.

\subsection{Comparison}

This section presents the performance of two traditional \acs{ML} models, \acs{XGBoost} and \acs{KNN}. We choose XGBoost because it is often used in \ac{NID} studies \cite{xgboostids}, and it can perform well on the general tabular dataset. The \ac{KNN} is also included because it also learns by comparing sample distance in the dataset, which is relevant to the concept of \ac{CL}. Tbl.~\ref{tab:ml} shows the results on three dataset splits, including our best SwapCon model. We regard the best SwapCon model as the combination described in Tbl.~\ref{tab:numemb} with the \acs{PLE} numerical embedding strategy. 

\begin{table}[H]
  \centering
  \caption{Comparisons between SwapCon and ML models.}
  \label{tab:ml}
  \begin{tabular}{l|S|S|S}
    \toprule
    {AUC($\uparrow$)} & {SwapCon best} & {XGBoost} & {KNN} \\
      \midrule\midrule
    IID& \textbf{99.35} & 97.29 & 93.71 \\
    NEAR& \textbf{98.85} & 83.87 & 82.84\\
    FAR& \textbf{72.93} & 47.31 & 32.36 \\
    \bottomrule
  \end{tabular}
\end{table}

The results show that the two ML models are both not robust to feature distribution shifts. However, the \acs{XGBoost} model performs better than \acs{KNN} on all three splits.

\section{Conclusion}
\label{sec:conclude}

In this paper, we addressed the problem of feature distribution shift in \acf{NID} and proposed SwapCon, a solution using model pretraining. Our approach involves compressing time-invariant feature information into the model during the pretraining stage and refining the model for classification in the finetuning stage. We also explore the importance of numerical feature embedding in improving the accuracy of \acf{ML} models. We evaluated our approach on the Kyoto2006+ dataset and compared it with \ac{XGBoost} and \ac{KNN} based models. Our results show that pretraining with SwapCon increases robustness against feature distribution shift and outperforms the other \ac{ML} models in terms of accuracy. Pretraining brings the most benefit in the \textit{FAR} split where feature distribution shift affects the dataset the most. Our experiments also suggest that numerical feature embedding is an important factor in improving the performance of pretrained models. We also show that the model size impacts the pertaining gain. An in-depth study of the relationship between model size and pretraining gain remains for future work.


\bibliographystyle{IEEEtran}
\bibliography{sections/references}

\begin{acronym}
    \acro{IDS}[IDS]{Intrusion Detection System}
    \acro{HIDS}[HIDS]{Host-based Intrusion Detection System}
    \acro{NIDS}[NIDS]{Network-based Intrusion Detection System}
    \acro{NID}[NID]{Network Intrusion Detection}
    \acro{CV}[CV]{Computer Vision}
    \acro{NLP}[NLP]{Natural Language Processing}
    \acro{NCE}[NCE]{Noise Contrastive Estimation}
    \acro{LLM}[LLM]{Large Language Model}
    \acro{BERT}[BERT]{Bidirectional Encoder Representations from Transformers}
    \acro{MLM}[MLM]{Masked Language Model}
    \acro{NSP}[NSP]{Next Sentence Prediction}
    \acro{R2L}[R2L]{Remote-to-Local}
    \acro{U2R}[U2R]{User-to-Root}
    \acro{DL}[DL]{Deep Learning}
    \acro{ML}[ML]{Machine Learning}
    \acro{DT}[DT]{Decision Tree}
    \acro{SVM}[SVM]{Support Vector Machine}
    \acro{RNN}[RNN]{Recurrent Neural Network}
    \acro{PLE}[PLE]{Piecewise Linear Embedding}
    \acro{LE}[LE]{Learnable Embedding}
    \acro{t-SNE}[t-SNE]{t-Distributed Stochastic Neighbor Embedding}
    \acro{DNN}[DNN]{Deep Neural Network}
    \acro{ANN}[ANN]{Artificial Neural Network}
    \acro{CNN}[CNN]{Convolutional Neural Network} 
    \acro{AE}[AE]{AutoEncoder}
    \acro{TP}[TP]{True Positive}
    \acro{FP}[FP]{False Positive}
    \acro{TN}[TN]{True Negative}
    \acro{FN}[FN]{False Negative}
    \acro{AUC}[AUC]{Area Under the Curve}
    \acro{ROC}[ROC]{Receiver Operating Characteristic}
    \acro{PR-AUC}[PR-AUC]{Precision-Recall Area Under the Curve}
    \acro{TCP}[TCP]{Transmission Control Protocol}
    \acro{KNN}[KNN]{K-Nearest Neighbor}
    \acro{XGBoost}[XGBoost]{eXtreme Gradient Boosting}
    \acro{BCE}[BCE]{Binary Cross-Entropy}  
    \acro{CL}[CL]{Contrastive Learning}
    \acro{BN}[BN]{Batch Normalization}
    \acro{EB}[EB]{Exponential Binning}
\end{acronym}

\end{document}